\documentclass[runningheads]{llncs}
\usepackage{graphicx}
\usepackage{amssymb}
\usepackage{graphicx}
\usepackage{float}
\usepackage{lipsum}
\usepackage{booktabs,chemformula}
\usepackage[utf8]{inputenc}
\usepackage{subfig}
\usepackage{subfloat}
\begin{document}
\title{SeNA-CNN: Overcoming Catastrophic  \\ Forgetting in Convolutional Neural Networks \\ by Selective Network Augmentation\thanks{This work was supported by National Founding from the FCT- Fundação para a Ciência e a Tecnologia, through the UID/EEA/50008/2013 Project. The GTX Titan X used in this research was donated by the NVIDIA Corporation.}}
\titlerunning{SeNA-CNN}
%
\author{Abel Zacarias\inst{1}\orcidID{0000-0002-0226-9682} \and
Luís A. Alexandre\inst{1}\orcidID{0000-0002-5133-5025}}
\authorrunning{A. Zacarias and L. Alexandre}
%
\institute{Instituto de Telecomunicações,\\
Universidade da Beira Interior, Rua\\
Marquês d'Ávila e Bolama, 6201-001\\
Covilhã, Portugal\\
\email{\{abel.zacarias, luis.alexandre\}@ubi.pt}}
\maketitle              
\begin{abstract}
Lifelong learning aims to develop machine learning systems that can learn new tasks while preserving the performance on previous learned tasks. In this paper we present a method to overcome catastrophic forgetting on convolutional neural networks, that learns new tasks and preserves the performance on old tasks without accessing the data of the original model, by selective network augmentation. The  experiment results showed that SeNA-CNN, in some scenarios, outperforms the state-of-art Learning without Forgetting algorithm. Results also showed that in some situations it is better to use SeNA-CNN instead of training a neural network using isolated learning.

\keywords{Lifelong Learning, Catastrophic Forgetting, Convolutional Neural Networks, Supervised Learning.}
\end{abstract}
\section{Introduction}
Deep learning is a sub-field of machine learning which uses several learning algorithms to solve real-world tasks as image recognition, facial detection, signal processing, on supervised, unsupervised and reinforcement learning using feature representations at successively higher, more abstract layers. Even with the growth and success of deep learning on many applications, some issues still remain unsolved. One of these issues is the catastrophic forgetting problem \cite{2013arXiv1312.6211G}. This issue can be seen as an handicap to develop truly intelligent systems.

Catastrophic forgetting arises when a neural network is not capable of preserving the past learned task when learning a new task.
There are some approaches that benefit from previously learned information to improve performance of learning new information, for example fine-tuning \cite{6909475} where the parameters of the old tasks are adjusted for adapting to a new task and, as was shown in \cite{09601}, this method implies forgetting the old task while learning the new task. Other approach well known is feature extraction \cite{pmlr-v32-donahue14} where the parameters of the old network are unchanged and the parameters of the outputs of one or more layers are used to extract features for the new task. There is also a paradigm called joint train \cite{Caruana1997} where parameters of old and new tasks are jointly trained to minimize the loss in all tasks.

There are already some methods built to overcome the problem of catastrophic forgetting \cite{DBLPLiH16e}, \cite{MerrienboerBDSW15}, \cite{7280416}. But even with these and other approaches, the problem of catastrophic forgetting is still a big challenge for the Artificial Intelligence (AI) community and according to \cite{Silver13lifelongmachine} is now appropriate to the AI community to move toward algorithms that are capable of learning multiple problems over time.

In this paper we present a new method that is capable of preserving the previous learned task while learning a new tasks without requiring a training set with previous tasks data. This is achieved by selective network augmentation, where new nodes are added to an existing neural network trained on an original problem, to deal with the new tasks.  

SeNA-CNN is similar to progressive neural networks proposed in \cite{rusu-progressive-2016} and in the next section we present the main differences between the two methods.

This paper is structured as follows: section II presents related works on existing techniques to overcome the problem of catastrophic forgetting in neural networks. In section III we describe SeNA-CNN and some implementation details. Section IV presents the experiments and results of SeNA-CNN and on section V we present the conclusions.
\section{Related Work}
The problem of catastrophic forgetting is a big issue in machine learning and artificial intelligence if the goal is to build a system that learns through time, and is able to deal with more than a single problem. According to \cite{Bing}, without this capability we will not be able to build truly intelligent systems, we can only create models that solve isolated problems in a specific domain. There are some recent works that tried to overcome this problem, e.g., domain adaptation that uses the knowledge learned to solve one task and transfers it to help learning another, but those two tasks have to be related. This approach was used in \cite{DBLPJungJJK16} to avoid the problem of catastrophic forgetting. They used two properties to reduce the problem of catastrophic forgetting. The first properties was to keep the decision boundary unchanged and the second was that the feature extractor from the source data by the target network should be present in a position close to the features extracted from the source data by the source network. As was shown in the experiments, by keeping the decision boundaries unchanged new classes can not be learned and it is a drawback of this approach because it can only deal with related tasks, with the same number of classes, while in our approach, we are able to deal with unrelated problems with different number of classes.

The Learning without Forgetting (LwF) algorithm proposed in \cite{DBLPLiH16e} adds nodes to an existing network for a new task only in the fully connected layers and this approach demonstrated to preserve the performance on old tasks without accessing training data for the old tasks. We compare SeNA-CNN with LwF algorithm. The main difference is that instead of adding nodes in fully connected layers, we add convolutional and fully connected layers of the new tasks to an existing model and SeNA-CNN has a better capability of learning new problems than LwF because we train a series of convolutional and fully connected layers while LwF only trains the added nodes in the fully connected layer and hence, depends on the original task's learned feature extractors to represent the data from all problems to be learned. 

Progressive Neural Networks (PNN), proposed in \cite{rusu-progressive-2016}, also addressed the problem of catastrophic forgetting via lateral connection to a previous learned network. The main difference to SeNA-CNN is that the experiment was in reinforcement learning while our proposal is designed to work with supervised learning for image classification problems. This approach, as SeNA-CNN begins with one column, a CNN trained on a single problem. When adding new tasks parameters from the previous task are frozen and new columns are added and initialised from scratch. Another difference between PNN and SeNA-CNN, is that SeNA-CNN use the two first convolutional layers of the original model trained on isolated learning and by doing that SeNA-CNN can learn the new tasks faster than if all the layers had to be trained from scratch, while PNN adds an entire column each time that new tasks come and the new column is randomly initialised. In the experimental section \cite{rusu-progressive-2016} they demonstrated the proposed method with 2, 3 and 4 columns architecture on Atari Game and 3D maze game. For future work, as in our approach, the authors aims to solve the problem of adding the capability to automatically choose at which task a label belongs because during the experiment it was necessary on test time to choose which task to use for inference.

\section{Proposed Method}

Our proposal is a method that is able to preserve the performance on old tasks while learning new tasks, without seeing again the training data for old tasks, as is necessary in \cite{DBLPLiH16e}, using selective network augmentation. 

A model that is capable of learning two or more tasks has several advantages against that which only learns one task. First is that the previous learned task can help better and faster learning the new task. Second, the model that learns multiple tasks may result in more universal knowledge and it can be used as a key to learn new task domains \cite{ShinLKK17}.

\begin{figure*}%
    \centering
    \subfloat[Original model for old task trained on isolated learning.]{{\includegraphics[width=10cm, height=1.5cm]{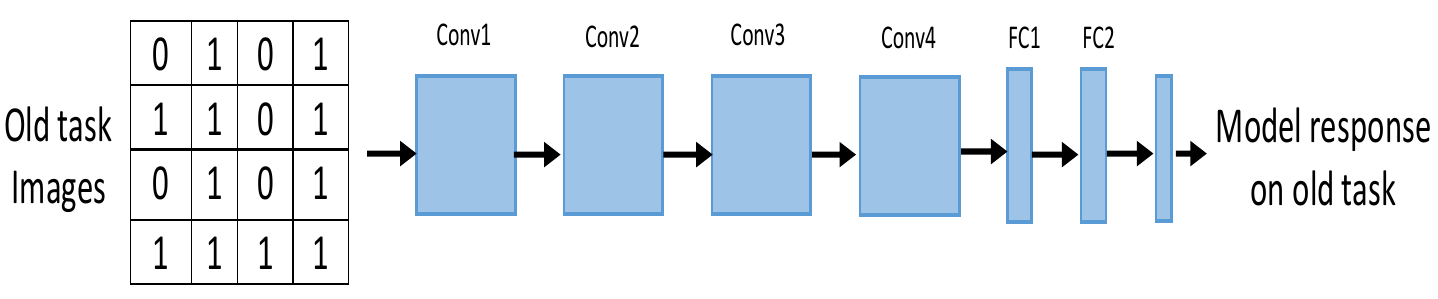} }}%
    \qquad
    \subfloat[Proposed model: adds new layers for the second task.]{{\includegraphics[width=10cm, height=4cm]{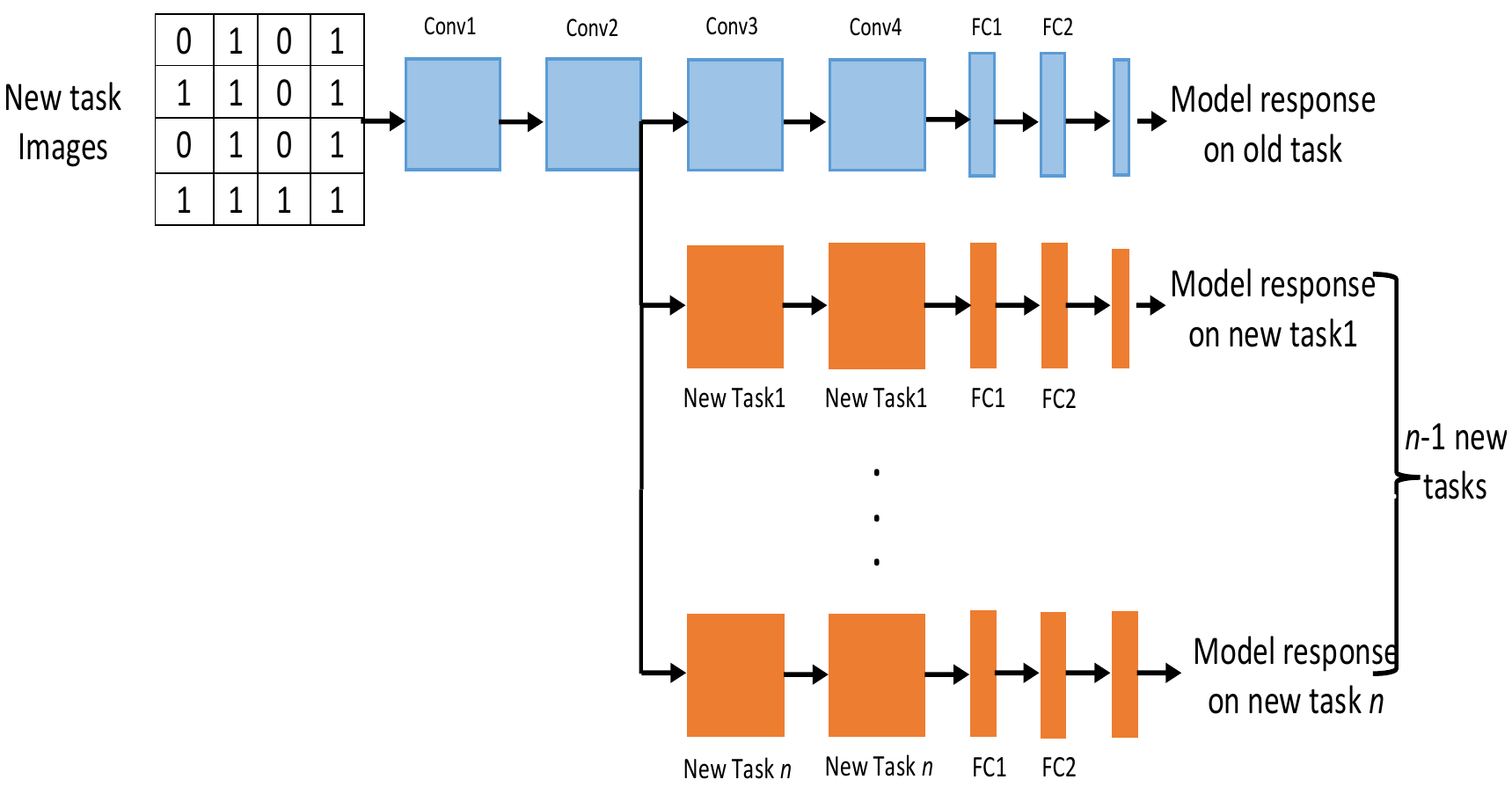} }}%
    \caption{Original and our model used in the experiment process to avoid the catastrophic forgetting by selective network augmentation. The blue coloured boxes correspond to the old task and the orange coloured correspond to the added layers.}%
    \label{fig}%
\end{figure*}

Initially a network is instantiated with \textit{L} layers with hidden layers \textit{h}$_{i}$ and parameters $\theta_{n}$ with random initialization. The network is then trained until convergence.
Fig. \ref{fig}(a) presents the original model for old task trained on isolated learning,  Fig. \ref{fig}(b) is our proposed model with two tasks. In Fig. \ref{fig}(b) the blue colour represents the old task network and the orange corresponds to the new added nodes for the new task. 

When a new tasks is going to be learned instead of adding nodes only in fully connected layers as is done in \cite{DBLPLiH16e}, we add layers for the new task Typically the added layers contain a structure similar to the network that we trained on  isolated learning. We consider the option of not adding the first two layers, because the neurons in those layers find several simple structures, such as oriented edges as demonstrated in \cite{RafegasVA17}. The remaining layers seem to be devoted to more complex objects, and hence, are more specific to each problem, and that is why we choose to create these new layers. It also resembles the idea of mini-columns in the brain \cite{doi:10.1152/jn.1957.20.4.408}. We add those layers and train them initialized with weights of old tasks, keeping the old task layers frozen.

When switching to a third task, we freeze the two previous learned tasks and only train the new added layers. This process can be generalized to any number of tasks that we wish to learn.

\section{Experiments}
We compared our method with the algorithm LwF proposed in \cite{DBLPLiH16e}. 

Our experiments evaluate if the proposed method can effectively avoid the catastrophic forgetting problem.
We conducted our experiments using three well known datasets namely CIFAR10 \cite{Acemoglu10mitand}, CIFAR100 \cite{Acemoglu10mitand} and SVHN2. Table \ref{tab1_1} shows information on each dataset, and the number of images on training and test sets. CIFAR10 AND CIFAR100 are very similar. CIFAR10 has 10 classes and these are subset of the 100 classes of CIFAR100. SVHN2 corresponds to street house numbers and has 11 classes. 

\begin{table}
   \centering
   \caption{Number of images for train and test sets.}
   \label{tab1_1}
   \begin{tabular}{@{}ccccc@{}}
     \toprule
     Data set & CIFAR10 & CIFAR100 &  SVHN2\\
     \midrule
     \ch{Train}  & 50000 & 50000 & 73257\\
     \ch{Test}  & 10000 & 10000& 26032\\
     \bottomrule
   \end{tabular}
 \end{table}
 
\begin{figure}%
    \centering
    {{\includegraphics[width=9cm, height=4.5cm]{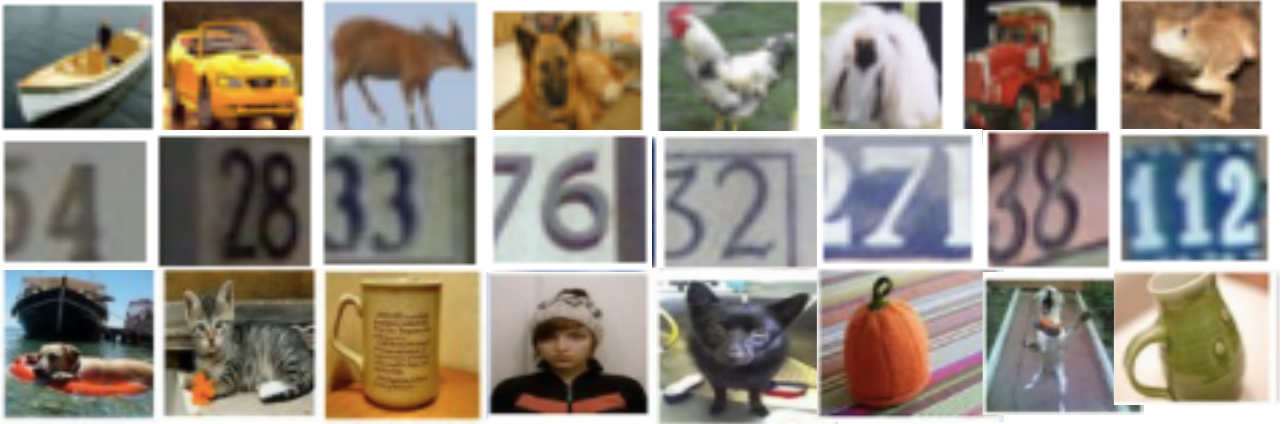} }}%
    \qquad
     \caption{Example images of the datasets used on the experiments. First row images corresponds to CIFAR10, second corresponds to SVHN2 and the last one are from CIFAR100 dataset.}%
    \label{fig:example}%
\end{figure}

\begin{figure}%
    \centering
    {{\includegraphics[width=9cm, height=5.7cm]{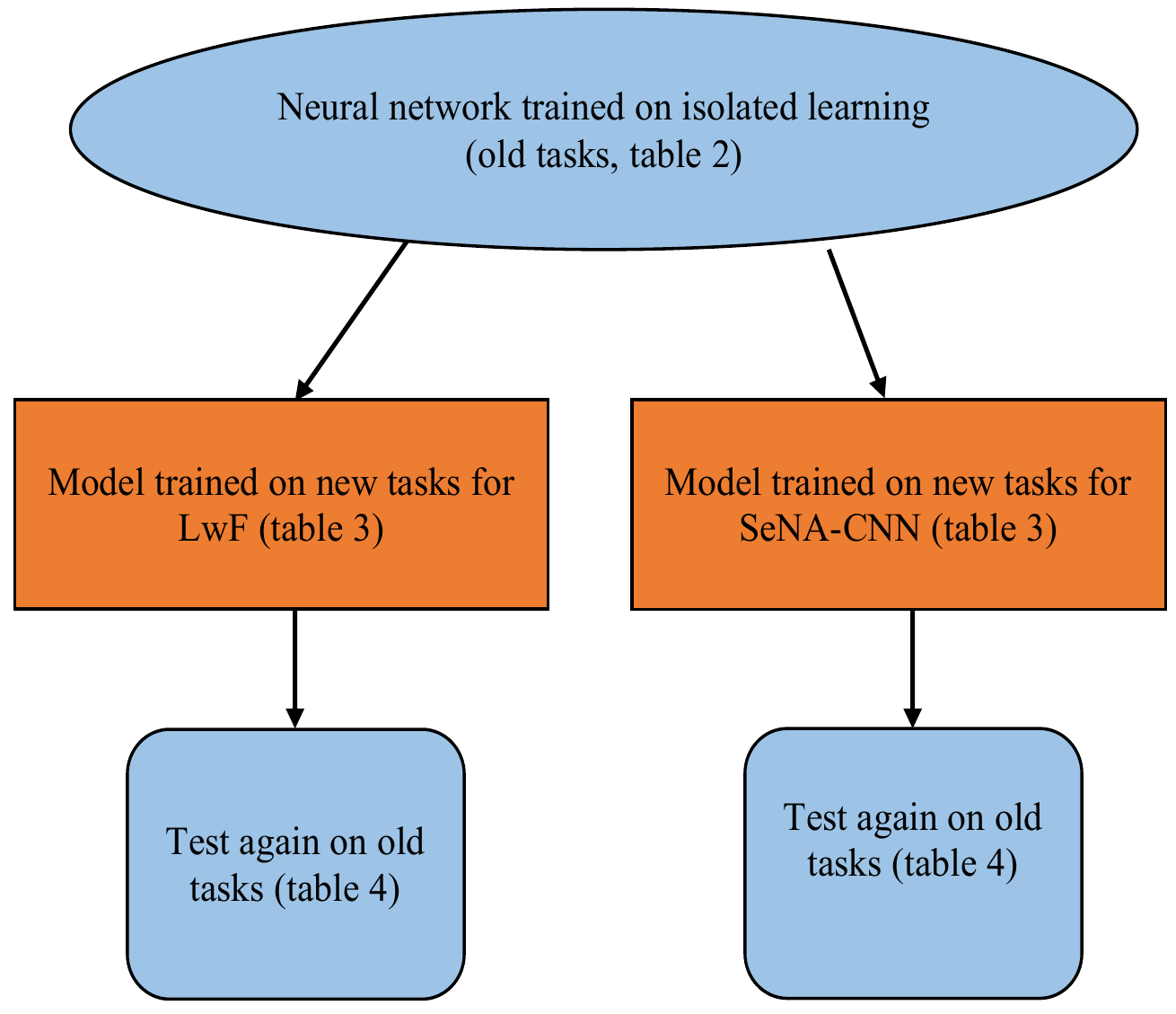} }}%
    \qquad
     \caption{Procedure used to test both evaluated models to overcome catastrophic forgetting.}%
    \label{Diagr}%
\end{figure}

Fig. \ref{Diagr} shows the procedure used to test the ability of both models (SeNA-CNN and LwF) to overcome catastrophic forgetting. Both models use the previous model trained on isolated learning. We add the new tasks and then evaluate the performance on the old tasks for each method.
\newpage
\subsection{Network Architecture}
We used a standard network architecture with an input layer followed by a convolution layer, a ReLU activation function, a convolution layer also followed by a ReLU, maxpooling and a dropout layer. The correspondent new task added to our model had two convolution layers, two ReLUs activation function and the last layer is the activation that corresponds to the softmax function with categorical cross-entropy loss, two dropout layers, a flatten layer and two dense layers one with 512 units and the other one corresponding to the number of classes for each task. At the test time the model predicts a single task and is necessary to choose which task to predict. At this point to predicts the desired task we set all other task labels to zero and only show to the model the one we want to predict. So far this process is done manually and we consider for future work the automatic choice of which task to predict.

Input images are RGB and have 32$\times$32 pixels. The first convolution layer has filters with dimensions 32$\times$32 while the other two convolution layers have filters with 64$\times$64. We used the keras API \cite{chollet2015keras} running on tensorflow \cite{tensorflow2015-whitepaper}.

\subsection{Training Methodology}

Our main goal is to evaluate if the proposed model learns new tasks while preserving the performance on old tasks. 
During training we followed the same practice as \cite{DBLPLiH16e}, the main difference is that we first freeze all layers of the original model and only train the added nodes. Then we train all weights for convergence using back-propagation with SGD algorithm with dropout enabled. All the networks had the same architecture, and the learning rate was set to 0.01, weight decay of 1e-6 and momentum 0.9. All networks use the same train, validation and test split for a given seed  number. Table \ref{tab2} shows the performance and execution time of each network after 12 training epochs. We run each experiment ten times and present results corresponding to the mean and standard deviation of these 10 repetitions. We run our experiments using a GeForce GTX TITAN X with 12 GiB.
\subsection{
Isolated Learning} We started by training 3 networks, one for each of the 3 data sets. Results of the experiment are shown in table \ref{tab2} where for each network we present the mean performance, its standard deviation and the execution time for train and test. These networks will be used both for SeNA-CNN and LwF in the next experiments.
 
\begin{table}
   \centering
  
   \caption{Network performance on isolated learning and execution time for train and test sets.}
    \label{tab2}
   \begin{tabular}{@{}ccccc@{}}
     \toprule
     Train & Test & Baseline [\%] & Execution Time [s]\\
     \midrule
     \ch{CIFAR}10  & CIFAR10 & 74.10$\pm0.70$ & 312 \\
     \ch{CIFAR}100  & CIFAR100 & 51.44$\pm0.40$ & 423\\
     \ch{SVHN}2  & SVHN2 & 92.27$\pm0.80$ & 438\\
     \bottomrule
   \end{tabular}
 \end{table}

\subsection{Adding New Tasks to the Models}

As Fig. \ref{Diagr} shows,we used the networks trained on isolated learning to implement our method by adding layers of the new tasks in such way that the model can learn a new task without forgetting the original one. Table \ref{two_tasks_Table1} presents the performance of the proposed method when adding new tasks and compares it with the baseline \cite{DBLPLiH16e}. These results correspond to the performance of our model and LwF when using a model trained on cifar10 for isolated learning and we added to the model as new tasks svhn2 and cifar100. This process was repeated for the other two tasks.

Results shows that SeNA-CNN outperformed LwF algorithm almost in all scenarios, showing that selectively adding layers to an existing model can preserve the performance on the old tasks when learning a new one, also is not necessary to train again the previous model and the new task learned will not interfere on the previous learned one. Overall SeNA-CNN outperformed LwF algorithm in 2/3 of the experiments showing the effectiveness of the proposed method to learn new tasks. 


\begin{table}
\centering
\caption{SeNA-CNN and LwF test accuracy (and standard deviation) on new tasks.}
\label{two_tasks_Table1}
\begin{tabular}{cccc}
\hline
Old  & New                 & LwF                & SeNA-CNN    \\ \hline
CIFAR10  & SVHN2& \textbf{84.02}(0.47)&82.27(0.38) \\ 
CIFAR10  & CIFAR100& 53.10(0.55)& \textbf{55.67}(0.52)\\
CIFAR100  & CIFAR10& 75.23(0.53)& \textbf{75.69}(0.52)\\
CIFAR100  & SVHN2&86.49(0.39) &\textbf{90.04}(0.38) \\
SVHN2  & CIFAR10& 66.42(0.62)&\textbf{67.27}(0.58) \\
SVHN2  & CIFAR100& \textbf{49.05}(0.63)& 47.15(0.45)\\
\hline 
\end{tabular}
\end{table}

We also evaluated if, when adding a new task, the knowledge previous learned was not overwritten. As shown in Fig. \ref{Diagr} we tested if the model was able to preserve the previous learned task. Table \ref{two_tasks_Table2} presents the results of these experiments. The second and third columns represent results of cifar10 as old task using the others two as new tasks. Similar setups are presented in the remaining columns. Results shows that our method outperformed LwF when remembering the previous learned tasks in all cases, and once again. We also verified that in some scenarios such as cifar100$\mapsto$cifar10 (for both methods), cifar100 performance increased compared to isolated learning, and it suggests using both proposed models instead of training from a random weights initialization, without interaction with other problems. These results are understandable since cifar10 and cifar100 are very similar and the two layers shared during the train of the new tasks increased the performance.        
Results show that by applying our method it is possible to overcome the problem of catastrophic forgetting when new tasks are added to the model.

\begin{table}
\centering
\caption{SeNA-CNN and LwF test accuracy (and standard deviation) showing that our  method does not forget old tasks after learning the new ones and outperforms the LwF method in all cases.}
\label{two_tasks_Table2}
\begin{tabular}{cccc}
\hline
New  & Old                 & LwF                & SeNA-CNN    \\ \hline
CIFAR10  & SVHN2& 87.96(0.75)&\textbf{89.84}(0.68) \\ 
CIFAR10  & CIFAR100& 52.39(0.43)& \textbf{53.34}(0.58)\\
CIFAR100  & CIFAR10& 69.37(0.65)& \textbf{70.59}(0.59)\\
CIFAR100  & SVHN2&89.01(0.39) &\textbf{89.53}(0.57) \\
SVHN2  & CIFAR10& 65.80(0.47)&\textbf{67.83}(0.59) \\
SVHN2  & CIFAR100& 48.11(0.41)& \textbf{49.40}(0.72)\\
\hline 
\end{tabular}
\end{table}
\subsection{Three Tasks Scenario} To demonstrate that SeNA-CNN is able to deal with several different problems, we experiment by learning three tasks. In this case we used the three datasets previously presented and we combine them two by two as old and one as new task. In Table \ref{three_tasks1} we presents results when adding a new task to a model that had already learned two tasks. From this scenario clearly in all cases SeNA-CNN outperformed LwF when learning a new task, and also the performance for cifar100 continue increasing for both methods and consolidating what we previously said.

\begin{table}
\centering
\caption{Three tasks SeNA-CNN and LwF test accuracy (and standard deviation) on new tasks.}
\label{three_tasks1}
\begin{tabular}{cccc}
\hline
Old  & New                 & LwF                & SeNA-CNN    \\ \hline
SVHN2, CIFAR10  & CIFAR100& 46.96(0.29)&\textbf{47.15}(0.48) \\ 
CIFAR10, CIFAR100  & SVHN2& 87.21(0.30)& \textbf{87.87}(0.50)\\
CIFAR100, SVHN2  & CIFAR10& 74.71(0.50)& \textbf{75.69}(0.14)\\
CIFAR10, SVHN2  & CIFAR100&54.24(0.37) &\textbf{54.87}(0.63) \\
SVHN2, CIFAR100  & CIFAR10& 65.99(0.47)&\textbf{66.00}(0.48) \\
CIFAR100, CIFAR10  & SVHN2& 87.68(0.43)& \textbf{89.08}(0.37)\\
\hline 
\end{tabular}
\end{table}

In this scenario we also evaluated the ability to preserve the performance of the two old learned tasks. Table \ref{three_tasks2} present results of both methods when they have to recall the old tasks. Comparing results, both algorithms typically had the same percentage of performance, meaning that in some scenarios SeNA-CNN performed better than LwF and vice-versa. Once again these results shows the ability to overcome the catastrophic forgetting problem in convolutional neural networks by selectively network augmentation.

\begin{table}
\centering
\caption{Three tasks SeNA-CNN and LwF test accuracy (and standard deviation) on old tasks.}
\label{three_tasks2}
\begin{tabular}{cccc}
\hline
New  & Old                 & LwF                & SeNA-CNN    \\ \hline
CIFAR100  & SVHN2, CIFAR10& \textbf{89.23}(0.70), 75.14(0.14)&89.01(0.44), \textbf{76.81}(0.64) \\ 
SVHN2  & CIFAR10, CIFAR100& \textbf{73.99}(0.12), \textbf{56.78}(0.37)&71.11(0.37), 56.20(0.58)\\
CIFAR10  & CIFAR100, SVHN2& \textbf{52.41}(0.26), 87.10(0.22)&49.14(0.58), \textbf{89.17}(0.57)\\
CIFAR100  & CIFAR10, SVHN2& 74.28(0.25), \textbf{90.04}(0.39)&\textbf{75.58}(0.52), 88.07(0.94) \\
CIFAR10  & SVHN2, CIFAR100& 90.13(0.59), \textbf{48.11}(0.27)&\textbf{90.19}(0.64), 46.96(0.51) \\
SVHN2  & CIFAR100, CIFAR10& 47.20(0.40), 74.95(0.43)&\textbf{47.87}(0.63), \textbf{75.24}(0.39)\\
\hline 
\end{tabular}
\end{table}
 
\section{Conclusion}
In this paper we presented a new method, SeNA-CNN to avoid the problem of catastrophic forgetting by selective network augmentation and the proposed method demonstrated to preserve the previous learned tasks without accessing the old task's data after the original training had been done. We demonstrated the effectiveness of SeNA-CNN to avoid catastrophic forgetting for image classification by running it on three different datasets and compared it with the baseline LwF algorithm. 

It has the advantage of being able to learn better new tasks than LwF since we train a series of convolutional and fully connected layers for each new task, whereas LwF only adds nodes to the fully connected layers and hence, depends on the original task's learned feature extractors to represent the data from all problems to be learned.

We also showed that in some scenarios SeNA-CNN and LWF increases the performance when compared to isolated training  for classification problems with some similarity. This is understandable since by reusing partial information from previous tasks, we are somehow doing fine-tuning on the  new task.

As future work we consider adapting SeNA-CNN for on-line learning and make it automatically choose which task is to be classified.

\bibliographystyle{splncs04}
\bibliography{Bibliografi}

\begin{thebibliography}{10}
\providecommand{\url}[1]{\texttt{#1}}
\providecommand{\urlprefix}{URL }
\providecommand{\doi}[1]{https://doi.org/#1}

\bibitem{tensorflow2015-whitepaper}
Abadi, M., Agarwal, A., Barham, P., Brevdo, E., Chen, Z., Citro, C., Corrado,
  G.S., Davis, A., Dean, J., Devin, M., Ghemawat, S., Goodfellow, I., Harp, A.,
  Irving, G., Isard, M., Jia, Y., Jozefowicz, R., Kaiser, L., Kudlur, M.,
  Levenberg, J., Man\'{e}, D., Monga, R., Moore, S., Murray, D., Olah, C.,
  Schuster, M., Shlens, J., Steiner, B., Sutskever, I., Talwar, K., Tucker, P.,
  Vanhoucke, V., Vasudevan, V., Vi\'{e}gas, F., Vinyals, O., Warden, P.,
  Wattenberg, M., Wicke, M., Yu, Y., Zheng, X.: {TensorFlow}: Large-scale
  machine learning on heterogeneous systems (2015),
  \url{https://www.tensorflow.org/}, software available from tensorflow.org

\bibitem{Acemoglu10mitand}
Acemoglu, D., Cao, D., Acemoglu, D., Cao, D.: Mit and cifar (2010)

\bibitem{09601}
Aljundi, R., Babiloni, F., Elhoseiny, M., Rohrbach, M., Tuytelaars, T.: Memory
  aware synapses: Learning what (not) to forget. CoRR  \textbf{abs/1711.09601}
  (2017), \url{http://arxiv.org/abs/1711.09601}

\bibitem{Caruana1997}
Caruana, R.: Multitask learning. Machine Learning  \textbf{28}(1),  41--75 (Jul
  1997). \doi{10.1023/A:1007379606734},
  \url{https://doi.org/10.1023/A:1007379606734}

\bibitem{chollet2015keras}
Chollet, F., et~al.: Keras (2015)

\bibitem{pmlr-v32-donahue14}
Donahue, J., Jia, Y., Vinyals, O., Hoffman, J., Zhang, N., Tzeng, E., Darrell,
  T.: Decaf: A deep convolutional activation feature for generic visual
  recognition. In: Xing, E.P., Jebara, T. (eds.) Proceedings of the 31st
  International Conference on Machine Learning. Proceedings of Machine Learning
  Research, vol.~32, pp. 647--655. PMLR, Bejing, China (22--24 Jun 2014),
  \url{http://proceedings.mlr.press/v32/donahue14.html}

\bibitem{6909475}
Girshick, R., Donahue, J., Darrell, T., Malik, J.: Rich feature hierarchies for
  accurate object detection and semantic segmentation. In: 2014 IEEE Conference
  on Computer Vision and Pattern Recognition. pp. 580--587 (June 2014).
  \doi{10.1109/CVPR.2014.81}

\bibitem{2013arXiv1312.6211G}
{Goodfellow}, I.J., {Mirza}, M., {Xiao}, D., {Courville}, A., {Bengio}, Y.: {An
  Empirical Investigation of Catastrophic Forgetting in Gradient-Based Neural
  Networks}. ArXiv e-prints  (Dec 2013)

\bibitem{7280416}
Gutstein, S., Stump, E.: Reduction of catastrophic forgetting with transfer
  learning and ternary output codes. In: 2015 International Joint Conference on
  Neural Networks (IJCNN). pp.~1--8 (July 2015).
  \doi{10.1109/IJCNN.2015.7280416}

\bibitem{DBLPJungJJK16}
Jung, H., Ju, J., Jung, M., Kim, J.: Less-forgetting learning in deep neural
  networks. CoRR  \textbf{abs/1607.00122} (2016),
  \url{http://arxiv.org/abs/1607.00122}

\bibitem{DBLPLiH16e}
Li, Z., Hoiem, D.: Learning without forgetting. CoRR  \textbf{abs/1606.09282}
  (2016), \url{http://arxiv.org/abs/1606.09282}

\bibitem{Bing}
Liu, B.: Lifelong machine learning: a paradigm for continuous learning.
  Frontiers of Computer Science pp.~1--3 (9 2016).
  \doi{10.1007/s11704-016-6903-6}

\bibitem{MerrienboerBDSW15}
van Merri{\"{e}}nboer, B., Bahdanau, D., Dumoulin, V., Serdyuk, D.,
  Warde{-}Farley, D., Chorowski, J., Bengio, Y.: Blocks and fuel: Frameworks
  for deep learning. CoRR  \textbf{abs/1506.00619} (2015),
  \url{http://arxiv.org/abs/1506.00619}

\bibitem{doi:10.1152/jn.1957.20.4.408}
Mountcastle, V.B.: Modality and topographic properties of single neurons of
  cat's somatic sensory cortex. Journal of Neurophysiology  \textbf{20}(4),
  408--434 (1957). \doi{10.1152/jn.1957.20.4.408},
  \url{https://doi.org/10.1152/jn.1957.20.4.408}, pMID: 13439410

\bibitem{RafegasVA17}
Rafegas, I., Vanrell, M., Alexandre, L.A.: Understanding trained cnns by
  indexing neuron selectivity. CoRR  \textbf{abs/1702.00382} (2017),
  \url{http://arxiv.org/abs/1702.00382}

\bibitem{rusu-progressive-2016}
Rusu, A.A., Rabinowitz, N.C., Desjardins, G., Soyer, H., Kirkpatrick, J.,
  Kavukcuoglu, K., Pascanu, R., Hadsell, R.: Progressive neural networks. arXiv
  preprint arXiv:1606.04671  (2016)

\bibitem{ShinLKK17}
Shin, H., Lee, J.K., Kim, J., Kim, J.: Continual learning with deep generative
  replay. CoRR  \textbf{abs/1705.08690} (2017),
  \url{http://arxiv.org/abs/1705.08690}

\bibitem{Silver13lifelongmachine}
Silver, D.L., Yang, Q., Li, L.: Lifelong machine learning systems: Beyond
  learning algorithms. In: in AAAI Spring Symposium Series (2013)

\end{thebibliography}
\end{document}